\pdfoutput=1

\documentclass[11pt]{article}

\usepackage[]{ACL2023}

\usepackage{amsmath}
\usepackage{amssymb}
\usepackage{mathtools}
\usepackage{amsthm}
\usepackage{array}
\usepackage{makecell}
\usepackage{array,multirow}
\usepackage{tabularx}
\usepackage{hyperref}
\usepackage{microtype}
\usepackage{graphicx}
\usepackage{booktabs}
\usepackage{algorithm}
\usepackage{algpseudocode}
\usepackage{times}
\usepackage{latexsym}
\usepackage[list=true]{subcaption}
\usepackage{microtype}
\usepackage{graphicx}
\usepackage{comment}
\usepackage[T1]{fontenc}

\usepackage[utf8]{inputenc}

\usepackage{microtype}

\usepackage{inconsolata}

\title{PESCO: Prompt-enhanced Self Contrastive Learning \\for Zero-shot Text Classification}

\author{Yau-Shian Wang \quad Ta-Chung Chi  \quad Ruohong Zhang \quad Yiming Yang \\
    Carnegie Mellon University \\
{\tt king6101@gmail.com  \{tachungc,ruohongz\}@andrew.cmu.edu} \\
{\tt yiming@cs.cmu.edu}
}

\begin{document}
\maketitle
\begin{abstract}
We present PESCO, a novel contrastive learning framework that substantially improves the performance of zero-shot text classification. We formulate text classification as a neural text matching problem where each document is treated as a query, and the system learns the mapping from each query to the relevant class labels by (1) adding prompts to enhance label matching, and (2) using retrieved labels to enrich the training set in a self-training loop of contrastive learning.  
PESCO achieves state-of-the-art performance on four benchmark text classification datasets. On DBpedia, we achieve 98.5\% accuracy without any labeled data, which is close to the fully-supervised result. Extensive experiments and analyses show all the components of PESCO are necessary for improving the performance of zero-shot text classification.

\end{abstract}


\section{Introduction}
Text classification is the task of assigning relevant category labels to each input document. It is an important problem in machine learning research with a wide spectrum of applications, including 
sentiment analysis~\cite{pang-etal-2002-thumbs,maas-etal-2011-learning, socher-etal-2013-recursive,tang-etal-2014-learning}, question answering~\cite{rajpurkar-etal-2016-squad,rajpurkar2018know}, and intent classification~\cite{5700816}, etc.
Recently, deep neural networks have obtained remarkable improvements in text classification, including  
CNNs~\cite{kim-2014-convolutional,NIPS2015_250cf8b5}, RNNs~\cite{tang-etal-2015-document,yang-etal-2016-hierarchical}, Transformers~\cite{NIPS2017_3f5ee243}, and more, thanks to the successful modeling of contextualized representations.

Despite the remarkable progress, training well-performing neural classifiers still requires a large amount of human-labeled documents, which is costly and time-consuming, especially for new application domains. 
This stimulates the recent trend of exploring self-supervised pre-training neural models on text classification tasks.
In particular, pre-trained language models (PTLMs)~\cite{devlin-etal-2019-bert,liu2019roberta,NEURIPS2019_dc6a7e65} clearly stand out from other methods owing to the pre-training on large-scale unlabeled data. 
%
Nevertheless, how to adapt PTLMs to downstream tasks with less supervision remains an open question for the research community, inviting new ideas to explore.

Prompt-based learning~\cite{NEURIPS2020_1457c0d6,shin-etal-2020-autoprompt,liu2021pretrain,li-liang-2021-prefix,gao-etal-2021-making} has been actively studied to 
better adapt PTLMs to downstream tasks with the goal of
reducing human annotation effort. For example,
PET~\cite{schick2020exploiting} is 
a prompt-based method for few-shot text classification. It formulates the task as a \textit{Cloze Test}, where a PTLM is used to predict the output label(s) by completing a prompt concatenated right after an input document.
For example, the sentiment of a product review is highly likely to be positive if a PTLM fills the word ``good'' into the following input:
\begin{quote}
    [Review] $\vert$ It is a \_ product.
\end{quote}
This example shows that prompt-based learning could unleash the potential power of a PTLM by constructing the input format of a downstream task in a way that closely resembles the PTLM pre-training objective, which is masked language modeling (MLM) in this case.


Motivated by the recent success of prompt-based learning, we propose PESCO, a novel self-training framework for zero-shot classification that uses prompts to enhance performance.
The self-training consists of two iterative steps, pseudo-label prediction and model update.
To make label descriptions more informative, we first put label descriptions into some predefined prompts and call the enhanced descriptions label-prompts.
As depicted in Figure~\ref{fig:formulation}, to predict the pseudo-label of a document, PESCO formulates text classification as a neural matching task.
A pre-trained text encoder maps both documents and label-prompts into a shared embedding space.
A label whose embedding is closest to the document is predicted as the pseudo-label.


To effectively update the text encoder with pseudo-labels, we propose the Prompt-enhanced Label-aware Cloze Test (PLCT), a contrastive learning framework for self-training.
The text encoder is trained to match a document and the text relevant to its pseudo-label.
The relevant texts include pseudo-label prompts and the key sentences from the documents assigned to the same pseudo-label.
The key sentence of each document is the sentence most related to its pseudo-label.

In our experiments, 
we show that the iterative self-training consistently improves the classification performance compared to the same model without self-training and that our proposed approach substantially outperforms other strong zero-shot classification baselines.
On some datasets, the zero-shot results are even on par with a fully supervised baseline.
On the Dbpedia dataset, in particular, PESCO achieves 98.5\% accuracy without any labeled data.

In summary, the contributions of this paper are twofold:
\begin{enumerate}
    \item 
    We explore text classification in a neural matching formulation enhanced by prompts. 
    We demonstrate that even without any finetuning on the text encoder, this straightforward formulation is an effective method for zero-shot text classification.
    
    \item
    The potential of contrastive learning for self-training has not been explored. We show that this is a promising direction for self-training and can achieve state-of-the-art performance on zero-shot text classification.
    
\end{enumerate}


\section{Related Work}

\subsection{Contrastive Learning}
Contrastive learning (CL)~\cite{1467314,10.1109/CVPR.2006.100} is a metric learning method that aims to pull closer similar inputs in the embedding space. 
Recently, the most popular and efficient methods for CL involve batch contrastive learning~\cite{he2019moco,chen2020simple}, which put similar inputs (positive pairs) and
dissimilar inputs (negative pairs) in the same
batch, simultaneously minimizing the distance
of representations from positive pairs, while maximizing the distance of negative pairs.

The key to CL is how to construct positive samples. Based on downstream applications, there are various ways to formulate the positive pairs. In self-supervised pre-training, the positive pairs are usually formulated by data augmentation. That is, different versions of a distorted sample are treated as a positive pair. 
In supervised contrastive learning~\cite{khosla2020supervised}, the examples belonging to the same class are viewed as a positive pair.

\begin{figure}
  \centering
  \includegraphics[width=7.42cm,height=3.5cm]{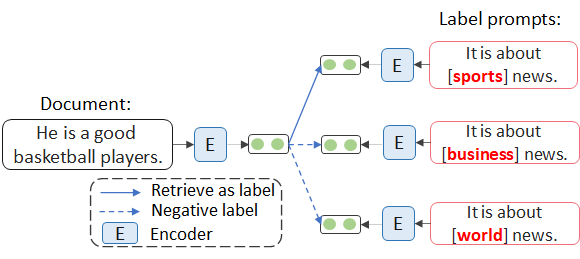}
  \caption{In this example, there are three classes, whose label descriptions are ``sports'', ``business'', and ``world'' respectively. We convert the descriptions into label-prompts by placing them into a template. The model predicts a label whose label-prompt embedding is the most similar to the document embedding.}
  \label{fig:formulation}
\end{figure}

In NLP, CL is usually used as an additional self-supervised pre-training to PTLMs because the sentence embeddings from PTLMs without fine-tuning are not ready to be used in downstream tasks~\cite{li-etal-2020-sentence}.
SimCSE~\cite{gao2021simcse} employs dropout as minimal data augmentation and obtains state-of-the-art unsupervised sentence representations.
In supervised SimCSE, the sentences with entailment relation are viewed as a positive pair.
Other approaches for data augmentation include sentence reformulation~\cite{Wu2020CLEARCL}, back translation~\cite{fang2020cert}, dual encoder~\cite{carlsson2021semantic}, language model corruption~\cite{Meng2021COCOLMCA}, and translation pairs~\cite{msimcse}.

In addition, CL is a commonly used training algorithm for neural text retrieval~\cite{xiong2021approximate}.
Inverse cloze test (ICT)~\cite{lee-etal-2019-latent} is the most commonly used contrastive pre-training task for retrieval that predicts a randomly selected sentence from the rest of the texts.
It is also possible to construct positive pairs by leveraging the document structures~\cite{Chang2020Pre-training}.

\subsection{Self-training and Zero-Shot Text Classifcation}
\paragraph{Self-training} Self-training~\cite{yarowsky-1995-unsupervised,10.1145/354756.354805,pseudo,9156610} is a widely used approach for semi-supervised learning and can have additive improvement to pre-training in both computer vision~\cite{NEURIPS2020_27e9661e} and NLP~\cite{du-etal-2021-self}.
The paradigm of self-training is first using a pre-trained base model as ``teacher'' to generate pseudo-labels on unlabeled data. 
The pseudo-label is then used to train a ``student'' model.
The teacher-student training is performed iteratively until convergence.

\paragraph{Zero-shot Text Classification}
Zero-shot classification aims to classify text using only label names without human annotation. Self-training has demonstrated impressive performance on few-shot~\cite{mukherjee-awadallah-2020-ust} and zero-shot text classification.
Unlike a few-shot setting which can use supervised information to obtain a base model, in zero-shot text classification, obtaining a base model is non-trivial.
LOTClass~\cite{meng-etal-2020-text} leverages PTLMs to augment label descriptions with semantically related words and then find category-indicative words among these related words to label documents.
They generalize the performance to the documents without category-indicative words via self-training.
iPET~\cite{schick2020exploiting} formulates text classification as a cloze test to help PTLMs understand the task.
They design several types of prompts for each dataset, and each type of prompt trains an individual teacher model to annotate documents using self-training.
A student model aggregates the knowledge from the teachers via knowledge distillation.
In this work, we propose a novel self-training method for zero-shot text classification that integrates self-supervised pre-training into self-training in a contrastive learning framework. 


\section{Zero-shot Classification as Matching} \label{sec:retrieval}
In our zero-shot setting, there are $N$ unlabeled documents $X=\{x_1,x_2,\cdots,x_N\}$ and a set of label descriptions $C=\{c_1,c_2,\cdots,c_L\}$, where $L$ denotes the number of classes.
We aim to learn a scoring function $g(x,c)$ so that relevant document and label description pairs can have higher scores.
A label whose label description has the highest score is selected as model prediction:
\begin{equation} \label{eq:predict}
    \hat{y} = \arg \max_{j} \  g(x,c_j),
\end{equation}

Inspired by the recent success of pre-trained sentence encoder~\cite{gao2021simcse,chuang-etal-2022-diffcse} which has shown impressive performance on matching relevant texts, we explore using pre-trained encoders as $g(x,c_j)$.
Specifically, as illustrated in Figure~\ref{fig:formulation}, we formulate zero-shot text classification as a neural text matching problem.
Both document and label descriptions are encoded into dense vectors by a shared encoder.
The matching score can be obtained by measuring cosine similarity between dense vectors.

However, label descriptions are usually a few words rather than a sentence with full semantics, which makes PTLMs unable to fully understand the meaning of the labels.
To tackle this, query reformulation~\cite{nogueira-cho-2017-task,petroni2020how} is a commonly used technique in retrieval to enhance the semantics of a query.
This technique can be further incorporated with prompt-based learning~\cite{schick2020exploiting}, which has shown that adding prompts to a text helps PTLMs understand classification tasks.
We use a prompt function $p(\cdot)$ to convert a label description $c$ into a prompt by placing label descriptions into pre-defined templates.
We design $T$ templates for each dataset, and the scoring function is:
\begin{equation} \label{eq:score}
    g(x,c) =  \frac{1}{T}\sum_{i=1}^{T} sim(f_{\theta}(x),f_{\theta}(p^i(c))),
\end{equation}
where $f_{\theta}(\cdot)$ is a text encoder with parameters $\theta$ that maps an input text to a dense embedding, and $sim(\cdot)$ is a similarity function.
For the rest of our paper, we use cosine similarity as $sim(\cdot)$.
For simplicity, in the rest of the article, we use $p_j$ to refer $p^i(c_j)$, which is the ``label-prompt`` of label $j$ with $i$ randomly sampled from $\{1,\cdots,T\}$.


\section{PESCO}

\begin{figure*}[th!]
  \centering
  \includegraphics[width=13cm,height=4.9cm]{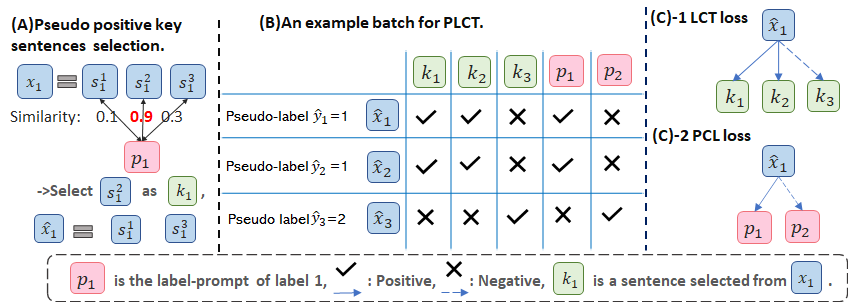}
  \caption{The framework of the PLCT. \textbf{(A)} Suppose the pseudo-label $\hat{y}_1$ for $x_1$ is $1$. We select $s^2_1$ as the key sentence $k_1$ for the document $x_1$ because the embedding of $s^2_1$ is the most similar to the embedding of label-prompt $p_1$. $\hat{x}_1$ is the augmented version of $x_1$, which removes $s^2_1$ from $x_1$.  \textbf{(B)} We use $k$ and $\hat{x}$ from part (A) to construct an example batch of PLCT with batch size $B=3$. Similar to self-supervised training, we use $\hat{x}_1$ to retrieve $k_1$ because they are from the same document. We use $\hat{x}_1$ to retrieve $k_2$ because $x_1$ and $x_2$ have the same pseudo-label. We also use $x_1$ to retrieve the its pseudo-label-prompt $p_1$. \textbf{(C)} We separate PLCT into LCT and PCL losses.}
  \label{fig:PLCT}
\end{figure*}

PESCO is a simple but effective self-training framework for zero-shot text classification.
Algorithm~\ref{alg:PESCO} gives an overview of PESCO.
In our iterative self-training loop, we first use a pre-trained sentence encoder $f_{\theta}$ to generate pseudo-labels (i.e. predicted labels) by the matching process described in Section~\ref{sec:retrieval}.
We then use the pseudo-labels to update $f_{\theta}$ by Prompt-enhanced Label-aware Cloze Test (PLCT), which leverages pseudo-labels to construct positive training pairs.
We continue the self-training process by iteratively generating pseudo-labels and updating the model using the PLCT objective function.

\subsection{Prompt-enhanced Label-aware Cloze Test} \label{sec:lct}
We propose Prompt-enhanced Label-aware Cloze Test (PLCT) to update our model using pseudo-labels.
As shown in Figure~\ref{fig:PLCT}, PLCT consists of two losses, Label-aware Cloze Test (LCT) loss and Prompt Contrastive Loss (PCL).
To compute LCT, for each document, we first select a key sentence from the document that is most relevant to its pseudo label.
In LCT, given a document, the positive texts are the key sentences from the documents belonging to the same pseudo-label.
For PCL, the positive texts for a document are its pseudo-label prompt (i.e. the label-prompt of a pseudo-label).
We combine these two losses by putting the positive texts of LCT and PCL into the same batch of a contrastive loss.

\subsubsection{Label-aware Cloze Test} \label{sec:LCT}
LCT is inspired by Inverse Cloze Test~\cite{lee-etal-2019-latent} which is a widely used self-supervised pre-training task for neural text retrieval.
It uses a randomly selected sentence from a document to match the remaining texts.
In a document, as some sentences don't contain useful information, using a randomly selected sentence for training is not an optimal choice. 
Instead, we use pseudo-label to select the key sentences.
Note that we use ``Cloze Test`` without ``Inverse`` because we use the remaining long texts to match its relevant short sentences, which can be viewed as label descriptions.

As illustrated in Figure~\ref{fig:PLCT}-(A), given an input document $x_i=\{s_i^1,s_i^2,\cdots,s_i^n\}$ consists of $n$ sentences and its predicted pseudo label $\hat{y}_i$, its key sentence $k_i$ is $s_j$, where:
\begin{equation} \label{eq:select}
    j = \arg \max_{n} \  g(s_i^n,p_{\hat{y}_i}).
\end{equation}
Here, $g(\cdot)$ is the scoring function in Eq.(\ref{eq:predict}).
As key sentence $k_i$ is more relevant to the pseudo-label than any other sentences in $x_i$, optimizing this objective is similar to minimize the distance between a document and its pseudo-label in embedding space, so $k_i$ can be viewed as an augmented version of the pseudo-label prompt.
Predicting the augmented version can have additional training signal than simply predicting pseudo-label prompt.
We provide a real example of $\hat{x}$ and $k$ in Table.~\ref{tab:example1} and more examples can be found in the Appendix Table~\ref{tab:example2}.

Since key sentences are highly correlated to corresponding pseudo-label prompts, given a document, it should not only match its key sentence but also key sentences in documents assigned to the same pseudo-label as shown in Figure~\ref{fig:PLCT} (C)-1.
We use the supervised contrastive loss~\cite{khosla2020supervised} to optimize LCT, which extends the SimCLR~\cite{chen2020simple} to allow multiple positive keys for a query in a supervised setting.
Specifically, let $\mathit{I} = \{1,\cdots,B\}$ be the set of the indices of the texts in a batch, where $B$ denotes the batch size.
The LCT loss $\mathcal{L}_{LCT}$ is written as:
\begin{equation} \label{eq:LCT}
 \sum_{i \in \mathit{I}} \frac{\scriptstyle -1}{\scriptstyle |K(i)|} \sum_{\hat{k} \in K(i)}\log{\frac{e^{sim(f_{\theta}(\hat{x}_i),f_{\theta}(\hat{k}) )/\gamma}}{\sum_{j \in \mathit{I}}e^{sim(f_{\theta}(\hat{x}_i),f_{\theta}(k_j) )/\gamma}}}.
\end{equation}
Here, $K(i) \equiv  \{k_j , \forall j \in \mathit{I} :\hat{y}_j=\hat{y}_i\}$ denotes the keys belonging to the same pseudo class $\hat{y}_i$, and $\gamma$ denotes a temperature commonly-used in CL.
To prevent trivial training signal, the input document is $\hat{x}_i=x_i \setminus \{k_i\}$ rather than $x_i$, where the key sentence $k_i$ is removed.

\begin{table*}[t!]
\small
\begin{tabularx}{\textwidth}{c|X|c}
\hline
Label Description & $x$ & $k$\\
\hline
\makecell{Family and\\ Relationship}  & \textcolor{blue}{how do you know if you're in love?} is it possible to know for sure? in my experience you just know. it's a long term feeling of always wanting to share each new experience with the other person in order to make them happy, to laugh or to know what they think about it. it's jonesing to call even though you just got off an hour long phone call with them. it's knowing that being with them makes you a better person. it's all of the above and much more. & \makecell{how do you know if\\ you're in love?} \\
\hline
\end{tabularx}
\caption{An example of the document $\hat{x}$ and the selected pseudo positive keys $k$ in Yahoo Answers. In this example, $k$ is very related to label description.}
\label{tab:example1}
\end{table*}


\subsubsection{Prompt Contrastive Loss} \label{sec:PLCT}
As the update target of self-training is to maximize the similarity between $x_i$ and its pseudo-label-prompt $p_{\hat{y_i}}$ in embedding space, we use the prompt contrastive loss (PCL) $\mathcal{L}_{PCL}$ to directly maximize the similarity:

\begin{equation} \label{eq:PCL}
    \mathcal{L}_{PCL}=-\sum_{i \in \mathit{I}} \log{\frac{e^{sim(f_{\theta}(\hat{x}_i),f_{\theta}(p_{\hat{y_i}}) )/\gamma}}{\sum_{c \in C}e^{sim(f_{\theta}(\hat{x}_i),f_{\theta}(p(c)) )/\gamma}}}.
\end{equation}
Depicted in Figure~\ref{fig:PLCT} (C)-2, this loss predicts $\hat{y_i}$ from $\hat{x}_i$.

\subsection{Combining LCT and PCL}
Naturally, to combine LCT and PCL, the simplest way is to use $\mathcal{L}_{PCL}+\mathcal{L}_{LCT}$ as the final training loss.
However, we found that minimizing this loss has limited improvement over minimizing $\mathcal{L}_{LCT}$ or $\mathcal{L}_{PCL}$ alone.
As depicted in Figure~\ref{fig:PLCT} (B), we come up with a more effective approach that puts the positive texts from these two losses into the same batch.
By doing so, pseudo keys $k$ and pseudo prompt $p$ can serve as mutually challenging negative samples, thus enhancing the representative power through more difficult contrastive tasks.
In our experiment, this simple solution significantly improves the performance.

Specifically, we use $\hat{x}_i$ as a query to retrieve (1) the key $k_i$ from the same text $x_i$, (2) $K(i)$, the keys belonging to the same pseudo class $\hat{y}_i$, and (3) the positive pseudo-label-prompt $p_{\hat{y}_i}$. The PLCT loss $\mathcal{L}_{PLCT}$ is written as:
\begin{equation} \label{eq:PLCT}
    \sum_{i \in \mathit{I}} \frac{\scriptstyle -1}{\scriptstyle |A(i)|} \sum_{a \in A(i)}\log{\frac{e^{sim(f_{\theta}(\hat{x}_i),f_{\theta}(a) )/\gamma}}{\sum_{m \in M}e^{sim(f_{\theta}(\hat{x}_i),f_{\theta}(m) )/\gamma}}}
\end{equation}
Here, $A(i) \equiv K(i) \cup \{p_{\hat{y}_i}\}$ is the set of positive texts in the mini-batch for $x_i$, $M \equiv \{k_j, \forall j \in \mathit{I}\} \cup \{p_c, \forall c \in C \}$ denotes the set of all the candidate keys.

Interestingly, $\hat{x}_i$ can be viewed as a challenging data augmentation of $x_i$ for predicting pseudo-label prompt because it removes the most salient sentence from $x_i$.
A model can make a prediction simply based on one salient sentence, neglecting the information of remainder.
This data augmentation method forces the model to capture additional information.



\begin{algorithm}
  \caption{PESCO}
  \textbf{Require:} Unlabeled texts $X$, label descriptions $C$.\\
  \textbf{Initialization:} A pre-trained sentence encoder $f_{\theta}(\cdot)$.\\
  \textbf{Repeat until convergence:}
  \begin{enumerate}
    \item Use $f_{\theta}(\cdot)$ to generate hard pseudo-labels $\hat{y}$ with Eq.(\ref{eq:predict}) for all unlabeled texts without data augmentation.
    \item Sample $T_t$ training pairs $(x,\hat{y})$ from step 1 based on the pseudo-label predicted probability. Use these pairs to update the $\theta$ of $f_{\theta}(\cdot)$ that minimizes the $\mathcal{L}_{PLCT}$ in eq~\ref{eq:PLCT}.
    \item With a more powerful $f_{\theta}(\cdot)$, go back to step 1.
  \end{enumerate}
  \textbf{Output:} $f_{\theta}(\cdot)$
 \label{alg:PESCO}
\end{algorithm}

\begin{table} [h] \small
\centering
\begin{tabular}{ccc}
\toprule
Dataset & Class Number & Test Examples \\
\midrule
AG News & 4 & 7,600\\
DBPedia & 14 & 70,000\\
Yahoo Answers & 10 & 60,000\\
Amazon & 2 & 400,000\\
\bottomrule
\end{tabular}
\caption{Dataset statistics.}
\label{table:datasets}
\end{table}

\vspace{-1mm}
\subsection{Self-training} \label{sec:self}
Algorithm~\ref{alg:PESCO} describes PECOS self-training loop.
Our self-training algorithm is a simplified version of noisy student training~\cite{9156610} that a single model alternately serves as a student and a teacher.
The key idea of noisy student training is that the teacher uses clean data without data augmentation to generate pseudo-labels, while the student learns to predict the pseudo-label on augmented data.

We first use pre-trained sentence encoder to initialize $f_{\theta}(\cdot)$.
Then, in step 1, $f_{\theta}(\cdot)$ serves as a teacher to generate pseudo-labels from clean data $x$ as described in Section~\ref{sec:retrieval}.
In step 2, $f_{\theta}(\cdot)$ serves as a student that learns to increase the probability of predicting pseudo-labels by minimizing $\mathcal{L}_{PLCT}$.
Step 2 is a noisy student training because the model takes $\hat{x}$ as input rather than clean $x$.
The self-training repeats step 1 and step 2 until convergence.
We use $f_{\theta}(\cdot)$ from the last iteration as our final model.

In the algorithm, we set $T_t = d \cdot T_{t-1}$ that gradually increases $T$ until a threshold ${T}'$.
The probability of sampling a pseudo training pair is proportional to the normalized scores outputed by the score function, so a more confident pseudo training pair is more likely to be sampled.
When sampling pseudo training pairs, we found that it is important to keep the ratio of all the labels balanced.
If a class doesn't have enough instances to be sampled, then we upsample the class to keep it balanced.

\begin{table*}[t!]
\small
\begin{tabularx}{\textwidth}{cXX}
\hline
Datasets & Label Descriptions & \makecell{Prompts}\\
\hline
AG news & (1)World (2)Sports (3)Business (4)Technology and Science &   \makecell{(1)Category: [desc]  news. \\ (2)[desc] news.}\\
\hline
DBpedia & (1)company (2)school and university (3) artist (4)athlete (5)politics (6)means of transportation (7)building (8)river and mountain and lake (9)village (10)animal species (11)plant and tree (12)album (13)film (14)novel and publication and book &   \makecell{(1)Category: [desc]. \\ (2)It is about [desc].}\\
\hline
Yahoo Answers & (1)Society and Culture (2)Science and Mathematics (3)Health (4)Education and Reference (5)Computers and Internet (6)Sports (7) Business and Finance (8)Entertainment and Music (9)Family and Relationships (10)Politics and Government & \makecell{(1)Category: [desc]. \\ (2)It is about [desc].}\\
\hline
Amazon-review-P & bad, good &   \makecell{(1)It is a [desc] product. \\ (2)In summary, the product is [desc]}\\
\hline
\end{tabularx}
\caption{The label descriptions and their prompts. [desc] in the templates denotes the label descriptions.}
\label{tab:prompt}
\end{table*}

\section{Experiments}
\subsection{Experimental Setting}
\paragraph{Implementation Details}
Inspired by ~\citet{yin-etal-2019-benchmarking} who formulate zero-shot text classification as entailment prediction, we choose the version of SimCSE~\cite{gao2021simcse} pre-trained on natural language inference (NLI) task~\footnote{We choose the model named ``sup-simcse-bert-base-uncased'' at \url{https://github.com/princeton-nlp/SimCSE}.} as our text encoder for all datasets.
Our experiments have shown that sentence encoder fine-tuned on NLI performs better on zero-shot classification tasks.
We use the representation outputted by the last layer as our sentence representation.

Following supervised contrastive learning~\cite{khosla2020supervised}, the value of $\gamma$ in all equations is set to be $0.07$.
For the value of $d$ in the self-training section, we set it to be $2$ because we want the model to annotate unlabeled data slowly.
The details of other hyperparameters in the Appendix~\ref{sec:hyper}.

\paragraph{Datasets}
We conduct experiments on various text classification datasets:  (1)\textbf{AG News}: topic classification on news article. (2)\textbf{DBpedia}: Ontology classification on selected classes from DBpedia. (3)\textbf{Yahoo Answers}: question type classification. (4)\textbf{Amazon}: binary sentiment classification on Amazon product review.
The statistics of these dataset are listed in Table~\ref{table:datasets}.

We provide the label descriptions in Table~\ref{tab:prompt}. The label descriptions of Yahoo Answers and AG news are mainly from the original dataset, and the label description of DBpedia is mainly from LOTClass~\cite{meng-etal-2020-text}.

\subsection{Effect of Using Prompts} \label{sec:p_eng}

\begin{table*} [h] \small
\centering
\begin{tabular}{cccccccc}
\toprule
Id & Self-train & Methods &  AG News & DBpedia & Yahoo Answers & Amazon\\
\midrule
$[1]$ & No & SimCSE w/o prompt & 69.7 & 73.8 & 55.2 &  88.3\\
$[2]$ & No & SimCSE w/ prompt & 76.3 & 76.0 & 56.5 & 88.3\\
$[3]$ & No & PET & 79.4 & 75.2 & 56.4 & 87.1 \\
\midrule
$[4]$ & Yes & iPET & 86.0 & 85.2 & 68.2 &\textbf{95.2} \\
$[5]$ & Yes & LOTClass & 86.4 & 91.1 & -- & 91.6 \\
\midrule

$[6]$ & Yes & PESCO w/o prompt & 87.1 & 96.0 & 69.9 & 95.1\\
$[7]$ & Yes & PESCO & \textbf{89.6} & \textbf{98.5} & \textbf{71.1} & \textbf{95.2}\\
\midrule
$[8]$ & -- & Supervised & 94.2 & 99.3 & 77.3 & 97.1\\
\bottomrule
\end{tabular}
\caption{Test-set accuracy of zero-shot text classification methods. The Self-train column indicates whether a method performs self-training on unlabeled data.}
\label{table:main_results}
\end{table*}

We investigate whether supplementing the label description with the prompt can help the model better understand the meaning of the label, and thus improve the performance.
In Table~\ref{tab:prompt}, we provide the label descriptions and the prompts we use.
For each dataset, we manually design two prompts, where the '[desc]' in the templates is the label description.
For example, given a label description ``Health'', the prompting function  converts it into either ``It is about Health'' or ``Category: Health''.

Our experiments showed that the choice of prompts doesn't affect performance much as long as reasonable prompts are given. 
For example, in AG news, without self-training, the accuracy of using “Category: <label> news”, “This is about <label> news”, and “<label> news” are 76.4, 76.0, and 78.0 respectively.
Furthermore, our scoring function, as described in Eq.(\ref{eq:score}), combines the scores of different prompts, which further reduces the gap.
The performance gap among different prompts is less than 2\% without self-training and less than 1\% after self-training.

In Table~\ref{table:main_results}, we analyze the effect of using prompts on SimCSE without self-training. By comparing [1] with [2], we find that using prompts for retrieval improves the performance on most of the datasets, especially on AG News.
We find that without the word ``news'', the model can not understand the meaning of the class only with the description ``world''.
Using the prompt-enhanced SimCSE [2] as the initial base model provides a better start for self-training.
However, comparing with the performance gap of [1] and [2] in Table~\ref{table:main_results}, we observed that the gap between [6] and [7]  becomes smaller, which indicates that the effect of using prompts decreases after self-training.


\begin{table*} [h]
\small
\centering
\begin{tabular}{ccccccc}
\toprule
Id &  Methods &  AG News & DBpedia & Yahoo Answers &  Amazon\\
\midrule
$[1]$ & PESCO & 89.6 & 98.5 & 71.1 & 95.2\\
$[2]$ & PESCO - R & 87.0 & 97.1 & 69.1 & 95.0\\
$[3]$ & LCT & 88.0 & 89.1 & 69.6 & 94.3\\
$[4]$ & LCT - R & 80.7 & 86.9 & 68.6 & 93.3\\
$[5]$ & PCL & 87.8  & 89.4 & 68.7 & 95.1\\
$[6]$ & LCT+PCL & 88.2 & 97.0 & 69.8  & 95.2\\
$[7]$ & PESCO w/o aug & 87.8 & 96.7 & 68.6 & 93.5\\

\bottomrule
\end{tabular}
\caption{Contrastive losses of different methods. The methods end with ``-R'' means their pseudo positive key sentences are randomly selected instead of picking the most salient sentence.}
\label{table:cl}
\end{table*}

\subsection{Zero-shot Text Classification}
In Table~\ref{table:main_results}, we compare our results against two state-of-the-art zero-shot text classification baselines, LOTClass~\cite{meng-etal-2020-text} and iPET~\cite{schick2020exploiting}.
We select these two methods as our baselines because they both employ self-training for zero-shot classification.
In [1], [2], and [3], they do not employ self-training on unlabeled data, so the Self-train column is ``No''.
In [7], we report the best results over 5 runs on PESCO single model performance without an ensemble.
We also report the average, maximum, and minimum accuracy over 5 runs in Appendix Table~\ref{tab:5runs}. 
In [8], to see the gap between zero-shot and fully-supervised settings, we train a typical BERT~\cite{devlin-etal-2019-bert} classifier on a labeled training set. We jointly finetune BERT and a linear classifier on top of BERT [CLS] output layer.

\paragraph{Effect of Self-training}
First, by comparing [7] against [2] in Table~\ref{table:main_results}, we find that the proposed self-training framework significantly improves the performance by more than 10\% on average.
On DBpedia, self-training improves performance substantially by 20\%, and it even achieves 98.5\% accuracy.
This demonstrates that self-training is an effective method to enhance performance after general pre-training, closing the gap between fully supervised training.

\paragraph{Comparison against LOTClass} 
Comparing [7] PESCO against [5] LOTClass, PESCO significantly improves the zero-shot text classification performance on all datasets.
LOTClass leverages PTLMs to find the category-indicative words which are semantically related to label descriptions. 
The documents containing category-indicative words are classified as the corresponding category.
Our method uses a pre-trained sentence encoder to define the relevance between document and category, which is more effective and requires less human heuristics.

\paragraph{Comparison against iPET}
Our main baseline is [4] iPET, which uses [3] PET as a base model to generate initial pseudo-labels followed by a series of self-training steps.
We find that our base model [2] achieves similar performance with [3] on all datasets except Ag News, on which ours lags behind by 3\%.
The lesson here is that using text retrieval as a means of text classification gives a similar performance to that using cloze tests.
Next, our full model [7] is also better than [4] iPET on three datasets while achieving similar performance on the Amazon dataset, demonstrating the effectiveness of our method.
Also, we notice that PET requires a massive model ensemble (e.g. 15 models) to achieve the reported accuracy.
We run their code with a PvP ensemble without using various random seeds for ensembling.
Even with this simplified setting, iPET still needs far more disk space (1.4 GB vs 26 GB) and more training time than us in that we do not need to train various models for model ensembling in each self-training step.

Note that It is not feasible to test our method using Roberta-base/large because language models without SimCSE finetuning poorly capture the semantic meaning of texts in cosine similarity space and cannot be used for retrieval. On the other hand, simCSE is finetuned for sentence embeddings, making language models lose text generation ability. Because iPET and LOTClass require language models to generate tokens, using SimCSE-Roberta for iPET or LOTClass is also not feasible.

\subsection{Ablation Study and Analysis}
\paragraph{Comparison of different contrastive losses}
The results of different contrastive learning losses are shown in Table~\ref{table:cl}.
In the table, LCT means we only use $\mathcal{L}_{LCT}$ in Eq.(~\ref{eq:LCT}) to train our model, PCL means we use $\mathcal{L}_{PCL}$, and LCT+PCL means we sum the $\mathcal{L}_{LCT}$ and $\mathcal{L}_{PCL}$ as our loss function rather than using PLCT loss which puts keys and label-prompts in the same batch.
The methods end with ``-R'' means the pseudo positive sentences $k$ are randomly selected from the documents instead of picking the most salient sentences.

In LCT, although it doesn't explicitly minimize the distance between an input document and its predicted pseudo-label-prompt, optimizing this loss still obtains performance similar to PLC.
This implies the selected key sentences can serve as augmented version of label-prompts.

Furthermore, we analyze the difference in the performance between using randomly selected sentences and the most salient sentences.
By comparing [1] and [2], and [3] and [4], we can see that the model has a significant performance drop in predicting randomly selected sentences.
This demonstrates the importance of choosing a salient sentence as the training target.

Finally, to demonstrate the effectiveness of putting pseudo-label-prompts and key sentences in the same batch, we compare [1] against [6].
[1] yields better performance than [6], which implies using this more challenging contrastive task allows the model to learn more general representations.

\paragraph{Effect of Data Augmentation}
In Table~\ref{table:cl}, [7] PESCO w/o aug means we use $x_i$ as a query to retrieve its positive examples $A(i)$ instead of using $\hat{x}_i$ as a query.
Comparing [1] and [7], removing the most salient sentence from a document is an effective data augmentation method that can greatly improve performance.
This is consistent with previous literature ~\cite{9156610} that updating student models with noisy data is important in self-training.


\section{Conclusion}
This paper presents a novel approach to zero-shot text classification, which significantly improves the SOTA results on four benchmark datasets by formulating the classification task as a prompt-enhanced retrieval problem and by combining the strengths of pre-trained language models and contrastive learning over pseudo-labeled data in a self-training loop. Our experiments in comparison with representative baselines and ablation analysis show evidence for the effectiveness of the proposed approach.

\section{Limitations}
The main limitation of our method is that it heavily depends on the quality of the label description. 
If a label description does not precisely describe the meaning of the label, our method cannot work. 
For some classification tasks such as microaggression detection, their labels have abstract meaning that is difficult to be understood by pre-trained language models. 
Similarly, our method cannot work on the domain that is not covered by the pre-training corpora of language models, such as the medical domain.

Another limitation of our method is that PLCT loss cannot handle short texts. 
If a text consists of only one sentence, PLCT loss will no longer work because LCT requires a document to be more than one sentence.
In this case, PCL loss can still be used for self-training.

\bibliography{custom}
\bibliographystyle{acl_natbib}

\clearpage
\appendix

\begin{table} [h!]
\small
\centering
\begin{tabular}{ccccc}
\toprule
& AG News & DBpedia & Yahoo & Amazon\\
\midrule
avg & 88.7 & 96.9 & 70.5 & 94.3\\
max & 89.6 & 98.5 & 71.1 & 95.2\\
min & 87.7 & 96.1 & 70.0 & 93.9\\
\bottomrule
\end{tabular}
\caption{Average/minimum/maximum accuracy over 5 runs.}
\label{tab:5runs}
\end{table}

\begin{table*} [h!] \small
\centering
\begin{tabular}{ccccc}
\toprule
& AG News & DBpedia & Yahoo Answers & Amazon\\
\midrule
Learning rate & 1e-5 & 1e-5 & 5e-6 & 5e-6 \\
Document length & 156 & 128 & 192 & 128 \\
Batch size & 32 & 32 & 32 & 32\\
Epsilon & 1e-6 & 1e-8 & 1e-8 & 1e-8\\
${T}'$ & 0.2N & 0.5N & 0.1N & 0.1N\\
Epoch & 5 & 5 & 2 & 1\\

\bottomrule
\end{tabular}
\caption{Hyperparameters.}
\label{table:hyper}
\end{table*}

\section{Discussion}
\paragraph{Text Classification as neural text retrieval}
Formulating text classification as neural retrieval is straightforward but not widely explored by previous work.
In this work, we show that this formulation can also obtain good performance with a well-pre-trained sentence encoder.
The benefit of this formulation over cloze test is that we don't need to restrict the label description to only one word.
PET requires a carefully selected word (verbalizer) to represent each class.
If a classification task has hundreds or even more than thousands of categories, it is not feasible to manually select a word to represent each class.
Furthermore, if the meaning of a category in a classification task is too abstract or complex, we cannot simply represent it with a single word.
Our formulation allows the model to describe categories using sentences or even short texts and maybe a better choice for more challenging classification tasks.

\paragraph{Contrastive Learning for Self-training}
The effect of contrastive learning for self-training is not well-studied by previous work.
Contrastive learning obtains impressive results on unsupervised representation learning.
In a supervised setting, it is also robust to noisy labels and noisy data, and it also shows impressive performance on a few-shot classification.
Considering these good properties of contrastive learning, we believe contrastive learning is a promising direction for self-training and propose  PESCO to explore its potential on zero-shot text classification.

\section{Hyperparameters} \label{sec:hyper}

\begin{figure}[h!]
  \centering
  \includegraphics[width=6cm,height=3.5cm]{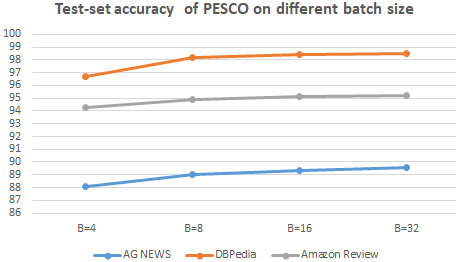}
  \caption{The effect of different batch sizes.}
  \label{fig:vis}
\end{figure}

\begin{figure}[h!]
  \centering
  \includegraphics[width=5cm,height=4cm]{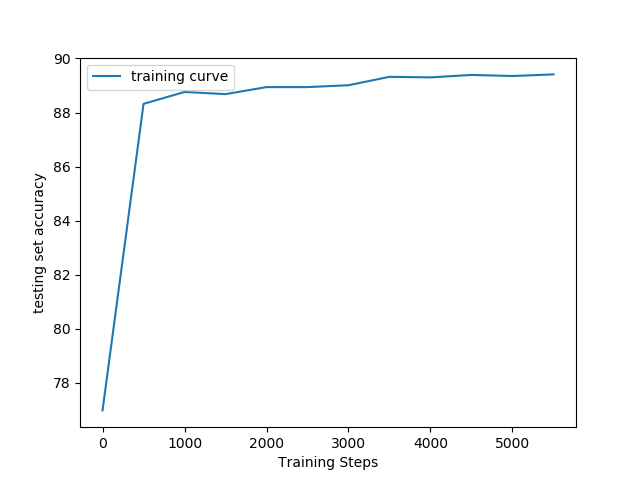}
  \caption{Training epoch versus validation set accuracy on AG News dataset.}
  \label{fig:training}
\end{figure}

As indicated by previous work~\cite{chen2020simple}, using a larger batch size generally yields better performance because it includes more negative samples.
We analyze how different batch size influences the performance of PESCO in Figure~\ref{fig:vis}.
We found that PESCO is not very sensitive to batch size. Using a smaller batch size only reduces the accuracy by less than 2\%.
Also, we observe that our algorithm converges after 1000 steps (1 epoch) of training, and additional training steps only slightly increase the performance. In other datasets, our algorithm also converges after 1 training epoch.

We list the hyperparameters of our model in Table~\ref{table:hyper}.
We use AdamW as our optimizer.
The ${T}'$ is the threshold mentioned in Section~\ref{sec:self}, we set it proportional to $N$, where $N$ is the total number of unlabeled data in the corresponding dataset.
We find that the number of training epoch only slightly influence the final performance that usually influences the accuracy by less than 1\%.
In Figure~\ref{fig:training}, we plot the training epoch versus validation set accuracy.
Although we train 5 epochs on the AG news to obtain the best result, the model actually converges in the early training stage.
Similar training curves can be observed in all the datasets.


\begin{table*}[t!]
\begin{tabularx}{\textwidth}{c|X|c}
\hline
Label Description & $x$ & $k$\\
\hline
Family and Relationship & \textcolor{blue}{where is the best place to look for love?} it might be easy to use the internet- there are many good matching web sites that can help & \makecell{where is the best place\\ to look for love?}\\
\hline
Entertainment and Music & \textcolor{blue}{what is the best place to get guitar lessons in the south bay area?} looking for a great instructor and relatively affordable price. i have no experience but have a desire to learn. it's really according to what you are looking for. certain teachers specialize in acoustic vs. electric (for example). your best bet is to place a request on a service such as click for lessons that will show you several teacher bios and let you decide for yourself. & \makecell{what is the best place\\ to get guitar lessons in \\the south bay area?}\\
\hline
Business and Finance & \textcolor{blue}{does anyone know a good apartment rental agency around washington dc?} i've had personal experience with archstone apartments and summit (just bought by camden) apartments in the past two years. while neither one is stellar, both were acceptable. both of these were in the northern virginia area - bedroom communities for d.c. best of luck apartment hunting! the housing market around here is absolutely insane. & \makecell{does anyone know a good\\ apartment rental agency around \\washington dc?}\\
\hline
Sports & \textcolor{blue}{why are there 5 rings in the olympics symbol?} what does it represent?   i heard few theories about it but not sure what is the correct one the 5 rings were introduced at the the 1920 games in antwerp games. the rings included at least one color from the flag of every participating country. & \makecell{why are there 5 rings\\ in the olympics symbol?} \\
\hline
\end{tabularx}
\caption{More examples of the distorted document $\hat{x}$ and the selected pseudo positive keys $k$ in Yahoo Answers. It happens that $k$ seems to be the most important sentence of the texts, so their semantics are closest to label descriptions.}
\label{tab:example2}
\end{table*}

\end{document}